%% file: main.tex
\documentclass[conference]{IEEEtran}

\IEEEoverridecommandlockouts                              

\usepackage{microtype}
\usepackage{amsmath,amssymb,amsfonts}
\usepackage{algorithmic}
\usepackage{graphicx}
\usepackage{textcomp}
\usepackage{xcolor}
\usepackage{subfig}
\usepackage{gensymb}
\usepackage{caption}
\usepackage{hyperref}
\usepackage{mwe}
\usepackage{amsfonts}
\usepackage{amssymb}
\usepackage{verbatim}
\usepackage{amsbsy}
\usepackage{siunitx}
\usepackage{tabularx, booktabs}
\usepackage{dblfloatfix}
\usepackage{cite}
\usepackage{cleveref}

\usepackage{algorithm, algorithmic}
\usepackage[autolanguage]{numprint}

\usepackage{spreadtab}
\usepackage{multirow}
\usepackage{cite}

\usepackage{hyperref}
\usepackage{xcolor}
\usepackage{algorithmic}
\usepackage{textcomp}
\usepackage{xcolor}
\usepackage{subfig}
\usepackage{gensymb}
\usepackage{caption}

\usepackage{mwe}
\usepackage{amsfonts}
\usepackage{amssymb}
\usepackage{verbatim}
\usepackage{amsbsy}
\usepackage{siunitx}
\usepackage{tabularx, booktabs}
\usepackage{soul}
\usepackage{tikz}
\usetikzlibrary{calc}
\usepackage{listings}

\makeatletter
\newif\if@anonymize

\@anonymizetrue    

\if@anonymize
  \newcommand{\highlight@DoHighlight}{
    \fill [outer sep = -15pt, inner sep = 0pt, color=black]
          ($(begin highlight)+(0,8pt)$) rectangle ($(end highlight)+(0,-3pt)$) ;
  }

  \newcommand{\highlight@BeginHighlight}{
    \coordinate (begin highlight) at (0,0) ;
  }

  \newcommand{\highlight@EndHighlight}{
    \coordinate (end highlight) at (0,0) ;
  }

  \newdimen\highlight@previous
  \newdimen\highlight@current
  \newlength{\item@width}

  \DeclareRobustCommand*\anonymize{%
    \SOUL@setup
    \def\SOUL@preamble{%
      \begin{tikzpicture}[overlay, remember picture]
        \highlight@BeginHighlight
        \highlight@EndHighlight
      \end{tikzpicture}%
    }%
    \def\SOUL@postamble{%
      \begin{tikzpicture}[overlay, remember picture]
        \highlight@EndHighlight
        \highlight@DoHighlight
      \end{tikzpicture}%
    }%
    \def\SOUL@everyhyphen{%
      \discretionary{%
        \SOUL@setkern\SOUL@hyphkern
        \SOUL@sethyphenchar
        \tikz[overlay, remember picture] \highlight@EndHighlight ;%
      }{%
      }{%
        \SOUL@setkern\SOUL@charkern
      }%
    }%
    \def\SOUL@everyexhyphen##1{%
      \SOUL@setkern\SOUL@hyphkern
      \settowidth{\item@width}{##1}%
      \makebox[\item@width]{}%
      \discretionary{%
        \tikz[overlay, remember picture] \highlight@EndHighlight ;%
      }{%
      }{%
        \SOUL@setkern\SOUL@charkern
      }%
    }%
    \def\SOUL@everysyllable{%
      \begin{tikzpicture}[overlay, remember picture]
        \path let \p0 = (begin highlight), \p1 = (0,0) in \pgfextra
          \global\highlight@previous=\y0
          \global\highlight@current =\y1
        \endpgfextra (0,0) ;
        \ifdim\highlight@current < \highlight@previous
          \highlight@DoHighlight
          \highlight@BeginHighlight
        \fi
      \end{tikzpicture}%
      \settowidth{\item@width}{\the\SOUL@syllable}%
      \makebox[\item@width]{}%
      \tikz[overlay, remember picture] \highlight@EndHighlight ;%
    }%
    \SOUL@
  }
\else
  \newcommand{\anonymize}[1]{#1}
\fi
\makeatother 

\title{\LARGE \bf
Reducing Latency in LLM-Based Natural Language \\ Commands Processing for Robot Navigation}

\author{Diego Pollini$^{1,2}$, Bruna V. Guterres$^{1}$, Rodrio S. Guerra$^{3}$ and Ricardo B. Grando$^{1}$
\thanks{$^{1}$Bruna V. Guterres, Ricardo B. Grando and Diego Pollini are with the Technological University of Uruguay. E-mail: {\tt\small ricardo.bedin@utec.edu.uy}}
\thanks{$^{2}$Diego Pollini is with National Technological University, Argentina. E-mail: {\tt\small diego.pollini@frra.utn.edu.ar}}
\thanks{$^{3}$Rodrigo S. Guerra are with the Federal University of Rio Grande. E-mail: {\tt\small rodrigo.guerra@furg.br }}
}

\begin{document}

\maketitle
\thispagestyle{empty}
\pagestyle{empty}

\input{sections/0_0_abstract.tex}
\vspace{-3mm}
\input{sections/0_1_supplementary_material.tex}
\vspace{-2.5mm}
\input{sections/1_introduction.tex}
\input{sections/2_related_works.tex}
\input{sections/3_methodology.tex}

\input{sections/4_experimental_results.tex}
\input{sections/6_conclusion.tex}

\input{sections/7_acknowledgment.tex}
\input{sections/8_references.tex}

\end{document}

%% file: sections/0_0_abstract.tex
\begin{abstract}
The integration of Large Language Models (LLMs), such as GPT, in industrial robotics enhances operational efficiency and human-robot collaboration. However, the computational complexity and size of these models often provide latency problems in request and response times. This study explores the integration of the ChatGPT natural language model with the Robot Operating System 2 (ROS 2) to mitigate interaction latency and improve robotic system control within a simulated Gazebo environment. We present an architecture that integrates these technologies without requiring a middleware transport platform, detailing how a simulated mobile robot responds to text and voice commands. Experimental results demonstrate that this integration improves execution speed, usability, and accessibility of the human-robot interaction by decreasing the communication latency by 7.01\% on average. Such improvements facilitate smoother, real-time robot operations, which are crucial for industrial automation and precision tasks.
  
\end{abstract}
\vspace{5mm}


%% file: sections/0_1_supplementary_material.tex


%% file: sections/1_introduction.tex
\section{Introduction}
\label{introduction}

The adoption of robotics in industrial applications has expanded significantly, driven by advances in computing hardware, sensor technology, and artificial intelligence (AI). Frameworks such as the Robot Operating System (ROS) have emerged as a foundational standard for industrial robotic applications, offering flexibility and extensive libraries to accelerate development cycles. ROS 2, the latest iteration, further enhances performance, security, and real-time capabilities, overcoming communication reliability and scalability limitations of its predecessor in distributed systems \cite{maruyama2016exploring}. Currently, advances in AI have given rise to new solutions for improvement in mobile robotics \cite{soori2023artificial,de2021soft} and sophisticated language models such as the GPT-3 and GPT-4 architectures having been adopted \cite{wake, liang}, capable of understanding context and generating coherent text. The integration of these models with ROS-based systems enables industrial environments to leverage intuitive natural language commands for enhanced flexibility and adaptability in robotic operations, broadening robotics' applicability through more effective human-robot interactions \cite{mao2023gpteval}.

\begin{figure*}[!ht]  
\centering \includegraphics[width=\linewidth]{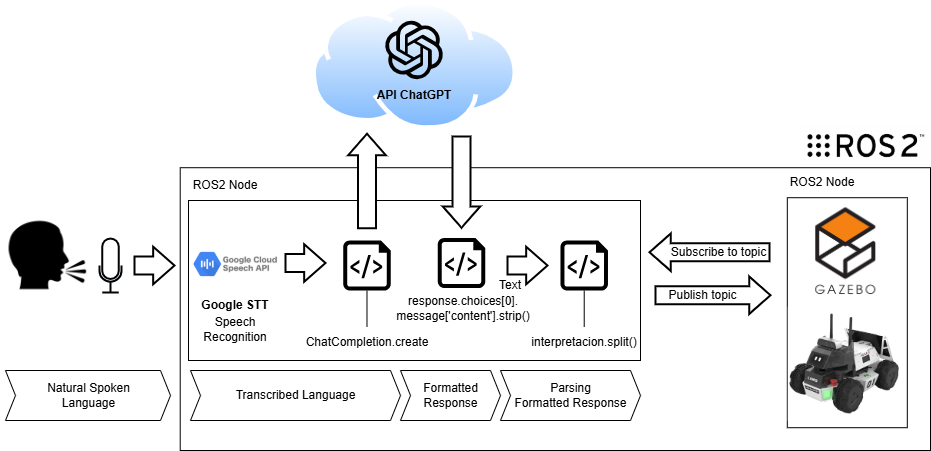}
\caption{Our proposed system.}
\label{figExemplo}
\end{figure*}

The purpose of this article is to explore and demonstrate the integration of ChatGPT with ROS 2. Specifically, we present an industrial environment that enables mobile robot navigation through natural language commands. The input prompt is sent to the GPT model, which interprets the command and generates properly formatted ROS 2 instructions for robot movement without requiring additional middleware. This integration demonstrates a more efficient approach that reduces latency in the command execution cycle, improving Human-Robot Interaction (HRI) for mobile robot navigation tasks. Figure \ref{figExemplo} illustrates our proposed system.

Key contributions of this work include:

\begin{itemize}

\item A ROS 2-based solution that interprets natural language voice commands and transforms them into control instructions for a mobile robot, without requiring rigid command syntax.

\item A direct integration architecture that enables human-robot interaction without intermediate middleware transport platforms such as JSON or Flask.

\item A comparative analysis between two GPT model versions, evaluating their performance metrics specifically for Human-Robot Interaction (HRI) navigation tasks with mobile robots.

\end{itemize}

This paper is organized as follows: Section~\ref{related_works} reviews related works. Section~\ref{methodology} describes our methodology, and Section~\ref{results} presents the experimental results. Finally, Section~\ref{conclusion} summarizes our contributions and discusses future work.

%% file: sections/2_related_works.tex
\section{Related Work}
\label{related_works}

Voice-based robotic control systems have long been an active research area. Many solutions rely on speech recognition technologies to interpret verbal commands and translate them into robot actions within industrial settings. The Google Voice Kit allows developers to build voice control interfaces using Google's speech recognition technology. This kit has been used to easily control robots using voice commands \cite{robotvoice1}. Furthermore, Amazon Alexa has been integrated into various robotic platforms to enable control using voice commands \cite{robotvoice2}.

AI-driven interaction with robotics has led to intuitive user interfaces, which allow more natural and effective interaction with robotic systems. However, these advanced interfaces typically do not use large language models (LLMs) like ChatGPT, but rather rely on task-specific speech recognition and command processing technologies. In this regard, Rasa is an open-source platform for building conversational assistants, including those that can control robots. It uses custom natural language processing (NLP) models to interpret voice and text commands. Wilcock et al. \cite{robotvoice3} show an implementation where a Furhat social robot uses Rasa conversational AI to access knowledge graphs in a Neo4j graph database. On the other hand, IBM Watson Assistant has been used in service robots to provide interaction capabilities using voice commands. This assistant uses NLP and machine learning technologies to understand and respond to user queries. For example, Di Nuovo et al. \cite{robotvoice4} implemented, on a SoftBank Robotics Pepper robot, IBM Watson Assistant to organize the workflow, Watson Text-to-Speech to generate the robot's voice, and Watson Speech-to-Text to perform speech recognition.

Unlike chatbots that only handle text, robotic systems require a deep understanding of real-world physics, environmental context, and the ability to execute physical actions. A generative robot model must possess strong common-sense knowledge and a sophisticated model of the world, as well as be able to interact with the user and the environment to reliably interpret and execute commands. These challenges are beyond the scope of traditional linguistic models, as they must not only understand the meaning of text but also translate intent into a logical sequence of physical actions. He et al. \cite {robotGPT} present the concept of RobotGPT as a multimodal approach to implementing the seven types of robotic intelligence that not only seek to incorporate various forms of intelligence, such as logical-mathematical and linguistic but also addresses the challenge of operating in a real-world environment, which is inherently complex and dynamic. By leveraging GPT's natural language processing capabilities, the author presents RobotGPT with the potential to improve human-robot interaction by facilitating a deeper understanding of context and enabling more accurate and adaptive responses to the situations at hand.

Wake et al. \cite{wake} presents an innovative approach to task planning in robotics by utilizing natural language models, specifically ChatGPT, in a few-shot learning context. This study introduces customizable prompts that allow complex human instructions to be translated into sequences of actions executable by robots, thus facilitating integration with robotic execution systems and visual recognition programs. One of the main contributions of the work is the ability to generate multi-step task plans that adapt to various environments, mitigating the limitations imposed by the number of tokens in ChatGPT. 

Liang et al. \cite{liang} introduces an innovative approach using large language models (LLMs) for generating robotic control programs, termed Language Model Programs (LMPs). This work focuses on the ability of LLMs to translate natural language instructions into executable code that can be used by robots in various tasks. The concept of "Code as Policies" (CaP) is introduced, which enables language models to generate programs that act as control policies for robots. 

Ahn et al. \cite{ahn} introduces a significant novelty at the intersection of language models and robotics by addressing the disconnect between high-level instructions and the execution of specific tasks by robots. The authors introduce the SayCan algorithm, which combines manipulation and navigation skills, allowing robots to interpret and execute natural language commands more effectively. One of the key contributions is the use of prompt engineering, which consists of carefully structuring inputs to guide the language model toward responses that are decomposed into low-level instructions suitable for robotic execution. 

Mirchandani et al. \cite{mirchandani} introduces a significant novelty by exploring how large language models (LLMs) can act as general pattern machines. Unlike traditional approaches that rely on domain-specific languages and program synthesis, this study demonstrates that LLMs can solve complex abstract pattern problems using in-context learning, where they are provided with input-output examples without the need for additional training or fine-tuning. 
The results indicate that LLMs can predict solutions for up to 85 out of 800 problems, outperforming some recent systems that rely on end-to-end machine learning methods. This finding suggests that LLMs are not only capable of manipulating linguistic patterns, but can also extrapolate non-linguistic patterns, opening up new possibilities for their application in reasoning and control tasks in robotics.

Xi et al. \cite{xi} presents technical developments in the field of large-scale language models (LLMs) and their application in the development of artificial intelligence agents. 
They introduce methods to improve memory efficiency by allowing agents to extract key details from historical interactions, including the use of prompts to concisely integrate memories and hierarchical methods that organize dialogues into both daily and general summaries. 

Hu et al. \cite{hu} presents a comprehensive review of foundation models applied to robotics, focusing on their ability to address robotic tasks in real-world environments and high-fidelity simulations. A foundation model is defined as a function that takes sensory inputs (such as images, textual descriptions, and audio signals) and context, to generate actions that fulfill specific robotic tasks. This unified approach allows the integration of multiple data modalities, which is crucial to improve the generalization and robustness of robots in diverse applications. Furthermore, the use of ROS (Robot Operating System) is highlighted as a fundamental platform that facilitates the implementation and development of these models, allowing the orchestration of different modules and the integration of sensory data. 

Hazra et al. \cite{hazra} presents an innovative combination of LLMs with a heuristic search-based planning framework, addressing limitations in plan generation in robotic environments. Unlike previous approaches that generated plans online and often resulted in infeasible actions, the authors perform offline planning, generating complete plans while maintaining an internal state of the world. This method uses precondition estimators and heuristics to improve the feasibility and cost-efficiency of the generated plans. 

Dipalo et al. \cite{dipalo} presents an innovative methodology called Keypoint Action Tokens (KAT), which enables pre-trained text-based Transformer models to act as few-sample imitation learning machines in robotics. Through KAT, visual observations and actions are transformed into sequences of tokens that can be interpreted by these models, facilitating the generalization of robotic skills from a limited number of demonstrations. The technique is based on in-context learning, where the model uses previous examples along with new inputs to predict actions without the need for additional parameter tuning. 

Yu et al. \cite{yu} introduces an innovative approach that combines LLMs with a low-level model prediction controller, MuJoCo MPC, to define reward functions in robotics. A system was developed that enables robots to perform complex manipulation and locomotion tasks, such as pushing objects and grabbing apples, using natural language instructions. The system translates these instructions into structured descriptions of robotic movements, improving stability and task-solving capability. It was implemented on a quadruped robot and a dexterous manipulator, achieving 90\% success in executing 17 tasks, compared to 50\% for a primitive skill-based approach. Furthermore, integration with ROS enables efficient communication between language models and robot controllers, facilitating real-time interaction management and adaptation of robotic movements. A specific regularization term was used to encourage stable movements in real-world environments, demonstrating the system's effectiveness in transferring from simulation to reality.

Along the same lines of integrating large language models (LLM) such as ChatGPT with robotic systems, Koubaa et al. \cite{koubaa} presents ROSGPT, a package developed for ROS2 that integrates ChatGPT with ROS2-based robotic systems. The implemented solution allows robots to interpret and execute natural language commands, significantly improving user experience and robot functionality. The integration of LLM with robotic systems, as demonstrated in Koubaa et al. \cite{koubaa}, represents a significant advance by enabling more sophisticated natural language understanding and generation, which improves human-robot interaction and expands the control and adaptability capabilities of robots.

Unlike previous works, our study investigates a direct API request model to enhance latency performance and minimize processing overhead, which is crucial in real-time industrial applications. The proposed system eliminates middleware dependency, processing responses from ChatGPT in structured lists without the need for JSON library, and improves execution speed essential for effective human-robot collaboration in industrial automation. In turn, the effectiveness of the GPT 3.5 and GPT 4.0 models is evaluated when interpreting the same unstructured commands from natural human speech and converting them into structured data that can be analyzed programmatically.

%% file: sections/3_methodology.tex
\section{Methodology}
\label{methodology}

The proposed architecture allows human-robot interaction based on the Agilex Limo mobile robot using natural spoken language. It enables the robot to perform linear movements or turns. Verbal commands are translated into text through the Whisper speech-to-text module and sent to ChatGPT using the OpenAI API. ChapGPT interprets the request and returns a formatted response that is parsed to produce ROS-compatible commands, allowing the robot to perform the action initially requested.

The script that manages the interaction between natural language and the control of the robot is implemented using the Python programming language. This choice is based on the versatility and wide adoption of Python in the robotics community, especially in the context of ROS 2. Python not only provides a clear and accessible syntax, but also has an extensive ecosystem of libraries and tools that facilitate the development and integration of complex systems. Thus, the ROS2 node that was developed facilitates the efficient integration of the communication logic with the OpenAI API and the Whisper speech-to-text module, also allowing the accurate generation of ROS commands for robot control. The hardware supporting the generated ROS2 node is a PC with sufficient capabilities to run the simulation on Gazebo and its microphone is used to receive the user's voice commands.

The system is started by configuring the OpenAI API key, which allows the use of the GPT model to interpret natural language commands. Communications with the robot is done with a publisher to send motion commands to the \textit{cmd\_vel} topic and a subscriber to receive data from the robot's IMU (Inertial Measurement Unit), thus updating the robot's yaw angle based on the received quaternion orientation.

A \textbf{record\_audio} function was made to record sound from a microphone and save the audio to a temporary WAV file, starting the recording when sound is detected and ending it after a period of silence. Subsequently, a \textbf{transcribe\_audio} function uses the Whisper model to convert the audio file into text.

Once the transcribed text is obtained, it is sent to the OpenAI API using a function that extracts and formats the commands necessary for controlling the robot. Depending on whether the command is of the move or rotate type, the corresponding messages are sent to the robot to perform the desired action.

Finally, the main loop of the system captures voice commands, interprets them, and executes the corresponding actions in the robot. The system is safely closed upon receiving an interrupt signal, ensuring proper termination of the ROS 2 node and resource cleanup.

Below is the pseudocode of the implemented script:

\begin{lstlisting}[linewidth=\columnwidth,breaklines=true,language=python,basicstyle=\footnotesize]

#OpenAI API Setup
    Read tOPENAI_API_KEY
    Set the API key
    
#Speech Recognition Setup
    Create a 'recognizer' object of `speech_recognition`
    Create a 'mic' object of `speech_recognition` 

#Defining the `RobotController` class
    `__init__` method
        Initialize the ROS 2 node
        Create a publisher for the 'cmd_vel'
        Create a subscriber for the 'imu/data'
        Initialize the 'yaw' variable to 0.0
    `imu_callback` method
        Update the 'yaw' variable from IMU
    `get_yaw_from_quaternion` method
        Calculate the 'yaw' angle
        Calculate the direction and time
        Publish a message to move the robot 
    `rotate` method
        Calculate the target angle and direction of rotation
        Publish to rotate the robot in the specified direction and angular speed
        Wait until the robot turns to the desired angle
        Stop the rotation

#Definition of the `recognize_vocal_order` function
    Capture Audio
    Transcribe audio to text using Google`s service
        If transcribed, return the text
        If error, return `None`

#Definition of the `interpret_sentence` function
    Create a message to the model with data to be interpreted
    Send the message to the model and receive the response
    Parse the response to move or rotate
    Return a tuple with the command and parameters 

#Definition of the `play_message` function
    Randomly choose a message from a list of options
    Use a TTS model to play the chosen message aloud

#Defining the `main` function
    Initialize the ROS 2 system
    Create an instance of the robot controller
    Run an infinite loop to:
        Call `recognize_vocal_command` to get a command from the user
        If a valid, interpret the command with `interpret_phrase`
            Execute command (move or rotate) 
        If not valid
            Print an error message
    Handle program interruption
    Shut down the ROS 2 system

\end{lstlisting}

\subsection{Human Robot Communication}

In the development of voice command robot control systems, the effective integration of speech recognition and synthesis is essential to achieve a natural, intuitive and efficient interaction. In this context, two key functions have been implemented. The first function is responsible for outputting text messages through a text-to-speech (TTS) engine, providing audible feedback to the user. This function not only informs the user about the current state of the system, such as starting to listen for commands, but also adds a layer of dynamism to the interaction by randomly selecting predefined messages. This approach ensures that communication is varied and less monotonous, improving the user experience. To achieve this, the function randomly selects a message from a predefined list of possible messages using the random module. Once the message is selected, the pyttsx3 TTS engine is used to convert the text into speech. 

\begin{lstlisting}[linewidth=\columnwidth,breaklines=true,language=python,basicstyle=\footnotesize]
def message_play():
    # Choose a random message
    message = random.choice(possible messages)  
    engine.say(message)
    engine.runAndWait()
\end{lstlisting}

The second function developed plays, captures, and transcribes the audio received through the microphone, using Google's speech recognition service to convert the audio into text. This function is initiated by setting the system microphone as the audio input source. Once the audio has been captured, the function proceeds to transcribe it into text using Google's speech recognition service. The captured audio is sent then to the Google service for processing. This allows the system to understand and process voice commands in the specified language.

The robustness of the function is ensured by efficient error handling. If the speech recognition service is unable to understand the audio due to problems such as excessive noise or unintelligible articulation, an exception is thrown. In such cases, the function prints a message indicating that the audio could not be understood and returns None. Also, if there is a problem connecting to the speech recognition service, such as network failures or problems in the Google service, an exception is thrown. The function handles this error by printing a message with the reason for the problem and also returns \textit{None}.

When transcription is successful, the function returns the recognized text, which is the textual representation of the captured vocal command.

\subsection{Voice Recognition}

Google speech recognition offers a convenient way to transcribe audio without the need to set up API keys, which is very useful for prototypes and small projects. In addition, it offers very good accuracy in transcribing speech to text, supporting multiple languages and accents. The user's speaking language and the language in which the transcription is intended must be configured. However, the es=ES model perfectly supports the English language. There are other models available in the library used that require an internet connection and an API key and support the Spanish language such as Microsoft Azure Speech, IBM Watson, Houndify that were not tested in this research work. The main reason was due to the limitations of computational resources compared to cloud processing, which uses servers with specialized and scalable hardware for handling deep learning models.

\subsection{Prompt Structure}

Once the transcript of the user's verbal command is obtained, a systematic process is initiated to interpret and execute that command using the ChatGPT model. The transcript of the verbal command is converted into a structured query directed to the ChatGPT language model through the OpenAI API. A function was created to organize this query into two distinct roles within the model: \textit{system role} and the \textit{user role}.

Properly formulating the prompts for the \textit{system role} and the \textit{user role} in the interaction with ChatGPT is crucial to obtaining accurate and structured responses that the script can efficiently parse and process. The \textit{system role} prompt sets the context and general instructions for the model, defining the role it should play and the rules it should follow. This prompt acts as a guide that directs ChatGPT to interpret and format responses consistently. On the other hand, the \textit{role user} prompt presents the user's specific request, in this case, the verbal instruction transformed into text. The clarity and precision in both prompts ensure that ChatGPT generates a structured response that the script can analyze to generate the ROS 2 language commands that set the robot in motion.

At the \textit{rol system} prompt, the model is set to act as a wizard for interpreting robot commands, providing clear instructions on converting magnitudes to standard units (meters and m/s) and assigning default values when the information is not available.

On the other hand, the \textit{rol user} prompt specifies the task of interpreting the given sentence and extracting the values needed to control the robot, with precise instructions on the expected response format, without quotation marks or a full stop. 

In this way, the model returns responses in a consistent and usable format. If the different ways the GPT model responds to different user questions are not considered, the script will not be able to publish the command to the corresponding ROS 2 topic that sets the mobile robot in motion.

Once ChatGPT has processed the query and generated a structured response, the script continues with the processing stage of this response, within the same \textit{interpret\_phrase} function, to convert it into executable commands in the ROS 2 ecosystem. The function starts by extracting the relevant content from ChatGPT's response, which is located in the content attribute of the response object. The text is cleaned by removing extra whitespace and ensuring that it is in a processable format.

\begin{lstlisting}[linewidth=\columnwidth,breaklines=true,language=python,basicstyle=\footnotesize]
    interpretation = response.choices[0].message['content'].strip()
\end{lstlisting}

\textit{response.choices[0]} accesses the first (and in this case, only) item in the choice list in the ChatGPT response. In the context of the OpenAI API, \textit{choices} is a list of responses generated by the model. Each object within \textit{choices} contains several fields. The \textit{message} field is one of these fields, and is a dictionary containing the structure of the generated response. The \textit{content} field of the \textit{message} dictionary stores the text string generated by the model. On the other hand, \textit{.strip()} is a string method that removes any whitespace at the beginning and end of the text string. This ensures that the response is clean and free of unwanted characters before and after the main content.

The \textit{interpretation} list is then split word by word using the \textit{.split} command. This step is critical to identify the individual components of the instruction based on the expected structure of the text generated by ChatGPT, such as command type, direction, distance, angle, and velocities.

\begin{lstlisting}[linewidth=\columnwidth,breaklines=true,language=python,basicstyle=\footnotesize]
    words = interpretation.split()
\end{lstlisting}

If the first word in the list is "move", the command is interpreted as moving the robot longitudinally forward or backward. The values for direction, meters, and speed are extracted. The meter's value is adjusted based on whether the direction is "backward", making the value negative if necessary.

\begin{lstlisting}[linewidth=\columnwidth,breaklines=true,language=python,basicstyle=\footnotesize]
if words[0] == "move":
    direction = words[1]
    meters = float(words[2])
    speed = float(words[6])
    if direction == "back":
        meters = -meters
    return ('move', meters, speed)
\end{lstlisting}

Conversely, if the first word is "rotate", the command is interpreted as being to rotate the robot. The values for direction, angle, and angular velocity are extracted. The angle is set to the specified direction (clockwise or counterclockwise).

\begin{lstlisting}[linewidth=\columnwidth,breaklines=true,language=python,basicstyle=\footnotesize]
elif words[0] == "rotate":
    direction = words[3]
    angle = float(words[4])
    angular_velocity = float(words[9])
    if direction == "clockwise":
        angle = -angle
    return ('rotate', angle, angular_velocity)
\end{lstlisting}

The \textit{interpret\_phrase} function returns a tuple composed of a string containing the type of action (move or rotate), a float containing the meters or the angle to rotate, and another float containing the value of linear speed or angular speed as appropriate.

\section{Commands in the Robot}

If the action to be performed is to move, the script calls the move method of the controller object (an instance of the \textit{RobotController} class). The move method receives two arguments: the distance in meters and the speed in meters per second, which are extracted from the tuple. The move method then creates a message of type \textit{Twist} and publishes it to the \textit{cmd\_vel} topic to tell the robot to move in a straight line at the specified speed for the time calculated based on the distance. If the order was to move forward, the direction variable gets the value 1, which means the speed is positive, and if, on the contrary, the order was to move backward, the direction variable gets the value -1, which means the linear speed in x will now be negative.

       

If the action to be performed is to rotate, the script calls the rotate method of the controller object. The rotate method receives two arguments: the angle in degrees and the angular velocity in radians per second, which are extracted from the tuple. The rotate method also creates a message of type \textit{Twist}, but this message configures the robot's z component to rotate in the specified direction.



        


%% file: sections/4_experimental_results.tex
\section{Experimental Results}
\label{results} 





Using the time library for Python, the duration of the verbal command and the response time of the Google speech recognition model are determined to find the requisition latency. To evaluate the effectiveness of the speech recognition system, a sample of 20 verbal commands was randonmly taken, which were generated in a silent environment and recorded in real time using the built-in microphone of a laptop. These verbal commands were transcribed and the transcription result is compared with the original verbal command to determine the success rate. 


To measure the response time of the OpenAI API, the \textit{rosgpt.py} script of the ROSGPT package is modified by including the \textit{time} library. Within the \textit{askGPT()} method of the \textit{ROSGPTProxy()} class, the time just before sending the request to ChatGPT is captured in a variable, then in another variable, the time just after receiving the response from ChatGPT is captured. Finally, in another variable, the difference is stored and the latency value is printed on the screen.

The latency time of the ROSGPT model is compared with the model of the developed solution with GPT3.5 Turbo and GPT4.0. It can be observed that for 16 out of 20 cases, the latency time of the developed solution is lower, as can be seen in Table \ref{tab:Performance}. The average latency time is reduced 7.01 \%, considering an average latency of 1.2865 s with our solution with GPT3.5 Turbo to 1.1804 s with ROSGPT\cite{kouba2023}. With both GPT3.5 Turbo and GPT4.0, 100\% of success rate was also achieved, as seen in Table \ref{tab:Performance}, while a total of 14 successes was achieved with ROSGPT \cite{kouba2023}. In this regard, it is worth remembering that the solution developed in this research work used a direct request methodology to the OpenAI API, unlike the ROSGPT model, which uses a Flask server.

On the other hand, the ROSGPT model incorrectly interpreted 13 of the orders, which led to failures in the execution of movements. Table \ref{tab:Performance} details these misinterpreted outputs and from them the following observations are made:
\begin{enumerate}
    \item The ROSGPT model does not recognize speeds or lengths expressed in units other than m/s or m, respectively. Therefore, the user must manually convert before executing the order. Unlike this, in the developed solution, the \textit{rol system} prompt and the \textit{rol user} prompt that hosts the user's request are designed in such a way that ChatGPT performs the unit conversion if necessary.
    
    \item The ROSGPT model does not recognize right turn or left turn indications as the developed solution model does.

\end{enumerate}

\begin{table}
    \centering
    \begin{tabular}{p{0.2cm} p{0.7cm} p{0.7cm} p{0.7cm} p{0.7cm} p{0.7cm} p{0.7cm} p{0.7cm} p{0.7cm}}
    \toprule
    {\color[HTML]{000000} \textbf{N°}} & {\color[HTML]{000000} \textbf{Latency GPT3.5 Turbo [s]}} & {\color[HTML]{000000} \textbf{Success GPT3.5 Turbo}} & {\color[HTML]{000000} \textbf{Latency GPT4.0 [s]}} & {\color[HTML]{000000} \textbf{Success GPT4.0}} & {\color[HTML]{000000} \textbf{Latency \cite{kouba2023} [s]}} & {\color[HTML]{000000} \textbf{Success \cite{kouba2023}}}\\ 
    \midrule
        1&1.05&OK&1.83&OK&1.15&OK\\
    \midrule
        2&0.97&OK&1.71&OK&1.66&OK\\
    \midrule
        3&0.91&OK&1.69&OK&1.19&OK\\
    \midrule
        4&1.08&OK&1.26&OK&1.42&OK\\
    \midrule
        5&0.77&OK&1.81&OK&1.45&OK\\
    \midrule
        6&0.88&OK&1.75&OK&1.32&OK\\
    \midrule
        7&0.61&OK&1.25&OK&1.27&FAIL\\
    \midrule
        8&3.62&OK&1.98&OK&1.01&FAIL\\
    \midrule
        9&1.03&OK&1.69&OK&1.31&OK\\
    \midrule
        10&1.09&OK&1.99&OK&1.13&FAIL\\
    \midrule
        11&1.46&OK&1.93&OK&1.46&OK\\
    \midrule
        12&0.72&OK&1.37&OK&1.37&OK\\
    \midrule
        13&0.97&OK&1.67&OK&1.61&OK\\
    \midrule
        14&0.72&OK&2.06&OK&1.37&OK\\
    \midrule
        15&0.92&OK&1.49&OK&1.29&OK\\
    \midrule
        16&0.83&OK&1.99&OK&1.37&OK\\
    \midrule
        17&1.36&OK&2.02&OK&1.26&FAIL\\
    \midrule
        18&0.88&OK&2.53&OK&0.99&FAIL\\
    \midrule
        19&0.79&OK&1.57&OK&1.03&OK\\
    \midrule
        20&1.27&OK&1.67&OK&1.07&FAIL\\  
    \midrule
    \multicolumn{8}{l}{\color[HTML]{000000} \textbf{Average Latency Time Ours with GPT3.5 Turbo: }{1.18}}\\
    \midrule
    \multicolumn{8}{l}{\color[HTML]{000000} \textbf{Average Latency Time Ours with GPT4.0: }{1.76}}\\
    \midrule
    \multicolumn{8}{l}{\color[HTML]{000000} \textbf{Average Latency Time \cite{kouba2023}: }{1.28}}\\
    \midrule
    \multicolumn{8}{l}{\color[HTML]{000000} \textbf{Total number of success cases Ours with GPT3.5 Turbo: }{20}} \\
    \midrule
    \multicolumn{8}{l}{\color[HTML]{000000} \textbf{Total number of success cases Ours with GPT4.0: }{20}} \\
    \midrule    
    \multicolumn{8}{l}{\color[HTML]{000000} \textbf{Total number of success cases \cite{kouba2023}: }{14}} \\
    \bottomrule
    \end{tabular}
    \caption{Performance Comparistion between approaches.}
    \label{tab:Performance}
\end{table}

%% file: sections/6_conclusion.tex
\section{Conclusions}
\label{conclusion}
The present work investigated the integration of ChatGPT with ROS2 for enhanced human-robot interaction in industrial settings by reducing latency and improving command interpretation. The implemented solution demonstrates an effective ability to transcribe, interpret, and execute voice commands in a robotic environment, providing a solid foundation for controlling a mobile robot within industrial settings using natural language, with a high success rate in correctly executing commands. An STT model has been successfully integrated with a GPT language model to interpret commands and generate instructions in the ROS 2 environment. The research has provided a detailed evaluation of the efficiency of Google's STT model and the GPT-3.5 Turbo and GPT-4.0 models, measuring response times and effectiveness in converting verbal commands to executable commands for ROS 2. The GPT-3.5 model outperformed the other investigated solutions in response latency and command interpretation accuracy, enabling more responsive and adaptable industrial robotic systems. This study supports automated workflows by enabling natural language interactions with industrial robots, minimizing manual interventions, and improving intuitive control. Future works include exploring faster speech-to-text models and leveraging vision-based systems for better situational awareness.



Alternatives could be explored to improve the speed of the STT model and the GPT model in order to achieve greater fluidity in the execution of the user's verbal orders. The use of other STT models in the cloud could be evaluated, which, a priori, could have lower latency than Google's STT, such as Deepgram or Microsoft Azure Speech-to-Text. Another option could be to implement a local model of OpenAI's Whisper. Although this would require having adequate computing infrastructure to support it and avoid processing delays (GPU, TPU), it would allow the simultaneous transcription and translation of the user's verbal command, which would enhance the performance of the system. Undoubtedly, the availability of better computing infrastructure would also improve the GPT model's response time. However, the use of Prompt Engineering techniques and tuning OpenAI API parameters such as temperature or the maximum number of tokens can help further improve latency.

%% file: sections/7_acknowledgment.tex
\section*{Acknowledgment}


The authors would like to thank the Technological University of Uruguay and the National Technological University of Argentina.


%% file: sections/8_references.tex
\bibliographystyle{./bibliography/IEEEtran}
\bibliography{./bibliography/main}